\documentclass[conference]{IEEEtran}
\IEEEoverridecommandlockouts

\usepackage{cite}
\usepackage{amsmath,amssymb,amsfonts}
\usepackage{algorithm}
\usepackage{algorithmic}
\usepackage{float}
\usepackage{graphicx}
\usepackage{textcomp}
\usepackage{xcolor}
\usepackage[hidelinks]{hyperref}
\usepackage[nameinlink,capitalise,noabbrev]{cleveref}
\crefname{figure}{Fig.}{Figs.}
\Crefname{figure}{Fig.}{Figs.}
\crefname{table}{Table}{Tables}
\Crefname{table}{Table}{Tables}
\crefname{equation}{Eq.}{Eqs.}
\Crefname{equation}{Eq.}{Eqs.}

\begin{document}
\raggedbottom

\title{Classifier-Dependent Benefits of Pseudo-Labeling for Semi-Supervised Android Malware Attribution}

\author{\IEEEauthorblockN{Md Rafid Islam, Zahid Hasan, and Hafiz Abdur Rahman$^*$}
\IEEEauthorblockA{\textit{Department of Electrical and Computer Engineering} \\
\textit{North South University}\\
Dhaka, Bangladesh \\
md.islam.241@northsouth.edu, zahid.hasan.241@northsouth.edu \\
$^*$hafiz.rahman@northsouth.edu (corresponding author)}
}

\maketitle
\thispagestyle{IEEEtitlepagestyle}  

\begin{abstract}
Detecting and classifying Android malware families remains challenging due to high feature dimensionality, class imbalance, and the high cost of expert-labeled data. Semi-supervised learning (SSL) offers a way to leverage unlabeled samples, but prior works rarely test whether SSL benefits generalize across classifier types or report statistical significance. We present a systematic evaluation of pseudo-labeling across six classifiers (LightGBM, XGBoost, Random Forest, Logistic Regression, MLP, and SVM) on the CICMalDroid 2020 dataset, using five-fold stratified cross-validation and paired t-tests across five labeled ratios (1--20\%). We find that SSL benefit is strongly classifier-dependent: SVM shows the largest significant gain ($+4.4\%$ accuracy at 5\% labels, $p = 0.0028$), LightGBM improves modestly ($+0.8$--$1.3\%$ at 2--5\% labels), while Random Forest is significantly harmed at low label ratios ($-3.1\%$ at 1\% labels). Per-class analysis reveals SSL disproportionately benefits the hardest-to-classify families, with Adware F1 improving by $+13.8$ percentage points versus only $+0.8$ for the already well-classified Benign class. We further show that approximately 800 labeled samples (10\% of the dataset) yield near-optimal performance across all classifiers. These findings offer practical guidance on when and with which classifier pseudo-labeling is worthwhile for Android malware classification.
\end{abstract}


\section{Introduction}

The complex evasion strategies used by modern malicious programs, such as code obfuscation, repackaging, and dynamic payload distribution, make traditional signature-based detection increasingly ineffective [1]. Differentiating between various malware families has become crucial for particular mitigating techniques, going beyond binary detection. High feature dimensionality, class imbalance, and most importantly, the lack of labeled data continue to make this multi-class classification problem difficult. Large datasets are costly and challenging to obtain since labeling Android applications requires expert analysis and substantial manual labor [2].

A promising approach is provided by semi-supervised learning (SSL), which uses a small labeled subset in addition to a large amount of unlabeled data. An initial model is trained on labeled data, confident predictions (pseudo-labels) are generated for unlabeled samples, and the model is then retrained on the combined set. Several studies have applied SSL to malware detection [3], but existing research has two critical gaps. First, the majority of research evaluates SSL on a single classifier, usually a gradient boosting model or neural network, without examining whether the benefits extend to other classifier families. Second, it is challenging to ascertain if reported improvements are real or the result of random variation because previous research seldom reports statistical significance or cross-validation variance. This is especially problematic because SSL techniques can be unstable, particularly when labeled data is scarce.

Through a methodical, controlled assessment of pseudo-labeling across six different classifiers (LightGBM, XGBoost, Random Forest, Logistic Regression, Multi-Layer Perceptron (MLP), and Support Vector Machine (SVM) with an RBF kernel), this paper addresses the gaps mentioned before. We use five-fold stratified cross-validation with paired t-tests for statistical significance to assess all classifiers at five labeled ratios (1\%, 2\%, 5\%, 10\%, and 20\% of total data) using the CICMalDroid 2020 dataset [4]. Our contributions are as follows:

\begin{itemize}
    \item We assess six classifiers that include tree-based, linear, kernel, and neural architectures under the same conditions and demonstrate that the SSL advantage is highly classifier-dependent.
    
    \item We report all results as mean $\pm$ standard deviation across five-fold cross-validation with paired t-tests ($\alpha = 0.05$), which provides confidence that observed effects are genuine.
    
    \item We show that SSL disproportionately helps the hardest-to-classify families---Adware improves by $+13.8\%$ F1 under SSL while already-well-classified Benign shows minimal gain ($+0.8\%$).
    
    \item We demonstrate that collecting approximately 800 labeled samples (10\% of the dataset) produces near-optimal performance (within 6--12\% of the fully supervised upper bound).
    
\end{itemize}

\section{Related Work}

The CICMalDroid 2020 dataset [4] is an established standard evaluation platform for Android malware research because of its variety and public accessibility. Mahdavifar et al. [4] trained pseudo-label deep neural networks (PLDNN) on dynamic behavioral profiles and achieved multi-class categorization accuracy of 96.7\% with an F1 score of approximately 94\%. This work demonstrated that semi-supervised techniques can be effective for unlabeled APKs and will reduce reliance on expensive manual annotation.

Several supervised machine learning studies have used CICMalDroid 2020 as their primary benchmark. Manzil and Naik [5] focused only on dynamic system call frequency features. They introduced a Huffman coding-based feature vector representation and achieved 98.70\% accuracy with a Random Forest classifier. Ansori et al. [6] applied gain ratio feature selection to reduce the feature space and then trained an ensemble of Random Forest, Extra Tree, and k-Nearest Neighbors classifiers that produced 94.57\% accuracy. More recently, Abedin and Mehrub [7] conducted a rigorous comparison of seven classifiers on CICMalDroid 2020 and found that XGBoost on the original (unreduced) features achieved the highest performance with 97.47\% accuracy and an F1 score of 97.16\%. Their experiments showed that dimensionality reduction with PCA consistently degraded all models. Deep learning approaches on CICMalDroid 2020 have also been explored by Usman et al. [8]. They combined BERT-based transfer learning with CNN ensembles for multi-modal classification. Yilmaz and Dogru [9] showed that converting static features to BERT-embedded images enables CNN classification that matches tabular methods without dynamic analysis.

Semi-supervised learning (SSL) for Android malware classification has been discussed in various works. For example, Chen et al. [10] proposed one of the earliest semi-supervised Android malware detectors, framing the problem as a Gaussian mixture model estimated via conditional expectation-maximization applied to dynamic API call logs. Sun et al. [11] addressed the high-dimensionality challenge inherent in Android feature spaces through a combination of random projection for dimensionality reduction with a confidence-guided SGD-based SVM. Muzaffar et al. [12] explored a lightweight semi-supervised path, trained a Naive Bayes classifier and achieved 93.256\% accuracy. Li et al. [13] applied contrastive learning to learn discriminative malware representations from unlabeled data. They achieved over 96\% accuracy for binary detection and over 98\% for multi-class family attribution.

While these works have established strong baselines on CICMalDroid 2020 and validated semi-supervised strategies for Android malware classification, none have systematically compared SSL benefits across multiple classifier families or reported statistical significance. This paper addresses these gaps by evaluating pseudo-labeling on six classifiers using five-fold cross-validation with paired t-tests, along with per-class analysis of SSL benefits.

\section{Methodology}

\subsection{Dataset Description}

We use the CICMalDroid 2020 dataset, which consists of 11598 samples and 470 dynamic features, divided into three categories: system calls, binder calls, and composite behavior. Table~\ref{tab:class-distribution} presents the complete class distribution.

\begin{table}[htbp]
\caption{Class-wise distribution and descriptions in the CICMalDroid 2020 dataset.}
\begin{center}
\begin{tabular}{l|c|p{4.5cm}}
\hline
\textbf{Category} & \textbf{Samples} & \textbf{Description} \\
\hline
Adware & 1253 & Displays intrusive ads or collects user data for advertising. \\
Banking & 2100 & Targets banking credentials or financial transactions. \\
SMS Malware & 3904 & Abuses SMS APIs to send messages or intercept codes. \\
Riskware & 2546 & Potentially harmful apps that pose security/privacy risks. \\
Benign & 1795 & Legitimate apps with no malicious behavior. \\
\hline
\end{tabular}
\label{tab:class-distribution}
\end{center}
\end{table}

All features are numerical and standardized to zero mean and unit variance using \texttt{StandardScaler} from scikit-learn. No feature selection or dimensionality reduction is applied to preserve the original feature space for fair comparison across classifiers.

\subsection{Experimental Setup}

\subsubsection{Train-Test Split}
The dataset is divided into a training pool (70\%, 8118 samples) and a stratified held-out test set (30\%, 3480 samples). To enable an objective assessment of the final model's performance, the test set is never utilized during model selection or hyperparameter tuning.

\subsubsection{Labeled Ratios}
To simulate realistic low-label scenarios, we sample labeled training sets at five different ratios of the total dataset: 1\%, 2\%, 5\%, 10\%, and 20\%. This corresponds to approximately 81, 162, 405, 811, and 1623 labeled samples, respectively. The remaining samples in the training pool (after extracting the labeled subset) are treated as unlabeled data for semi-supervised learning.

\subsubsection{Cross-Validation Protocol}
We use five-fold stratified cross-validation on the training pool for each labeled ratio to obtain reliable performance estimates. Stratification preserves the class distribution in every fold. Within each fold, we:

\begin{itemize}
    \item Sample the labeled set from the fold's training partition at the specified ratio (maintaining class balance)
    \item Train a supervised baseline on the labeled set
    \item Train a semi-supervised model using pseudo-labels generated from the unlabeled data
    \item Evaluate both models on the fold's validation set
\end{itemize}

This process is repeated for all five folds, which generate five independent accuracy measurements per model and labeled ratio.

\subsubsection{Evaluation Metrics}
We report two primary metrics: accuracy and macro F1-Score, which is the harmonic mean of precision and recall, averaged across classes without weighting by class size (appropriate for imbalanced data). Results are presented as mean $\pm$ standard deviation over five folds.

\subsubsection{Per-Class Analysis}
For the classifier and labeled ratio showing the strongest SSL improvement, we additionally perform per-class analysis using F1 scores computed individually for each malware family. This allows us to determine whether SSL benefits all classes uniformly or disproportionately affects specific families.

\subsection{Classifiers}

To determine whether the advantages of SSL rely on classifier architecture, we assess six classifiers from various model families. The classifiers and their important hyperparameters are compiled in Table~\ref{tab:classifiers}. To ensure that our results generalize rather than overfit this particular dataset, we purposefully forgo substantial hyperparameter tuning and use all classifiers with their default hyperparameters, unless otherwise specified.

\begin{table}[htbp]
\caption{Classifier configurations.}
\begin{center}
\resizebox{\columnwidth}{!}{%
\begin{tabular}{|l|l|l|}
\hline
\textbf{Classifier} & \textbf{Library} & \textbf{Key hyperparameters} \\
\hline
LightGBM (LGBM) & lightgbm & leaves=31, lr=0.05 \\
XGBoost (XGB) & xgboost & estimators=200, lr=0.05 \\
Random Forest (RF) & scikit-learn & estimators=100 \\
Logistic Regression (LR) & scikit-learn & max\_iter=1000 \\
MLP & scikit-learn & layers=(100,50), early stop \\
SVM & scikit-learn & RBF kernel, prob.=True \\
\hline
\end{tabular}%
}
\label{tab:classifiers}
\end{center}
\end{table}

\subsection{Semi-Supervised Learning: Pseudo-Labeling}

We use pseudo-labeling, a simple and widely used semi-supervised learning method that generates artificial labels for unlabeled data based on model confidence. Algorithm~\ref{alg:pseudo-labeling} presents the complete pseudo-labeling procedure used in our experiments.

\begin{algorithm}[t]
\caption{Pseudo-labeling for semi-supervised classification}
\label{alg:pseudo-labeling}
\begin{algorithmic}[1]
\REQUIRE Labeled set $\mathcal{D}_L = \{(x_i, y_i)\}_{i=1}^{N_L}$, unlabeled set $\mathcal{D}_U = \{x_j\}_{j=1}^{N_U}$, confidence threshold $\tau = 0.7$, number of classes $K = 5$
\ENSURE Final classifier $f_{\text{SSL}}$ trained on labeled and pseudo-labeled data
\STATE Train initial classifier $f_0$ on $\mathcal{D}_L$
\STATE Initialize empty pseudo-labeled set $\mathcal{D}_P \leftarrow \emptyset$
\FOR{each unlabeled sample $x_j \in \mathcal{D}_U$}
    \STATE Compute probability vector $p = f_0(x_j) \in [0,1]^K$
    \STATE Let $\hat{y}_j = \arg\max_c p_c$ \hfill \COMMENT{Predicted class}
    \STATE Let $\text{conf} = \max_c p_c$ \hfill \COMMENT{Prediction confidence}
    \IF{$\text{conf} \geq \tau$}
        \STATE Add $(x_j, \hat{y}_j)$ to $\mathcal{D}_P$ \hfill \COMMENT{High-confidence sample}
    \ENDIF
\ENDFOR
\STATE Combine labeled and pseudo-labeled sets: $\mathcal{D}_{\text{SSL}} \leftarrow \mathcal{D}_L \cup \mathcal{D}_P$
\STATE Train final classifier $f_{\text{SSL}}$ on $\mathcal{D}_{\text{SSL}}$
\STATE \textbf{return} $f_{\text{SSL}}$
\end{algorithmic}
\end{algorithm}

Since higher thresholds result in too few pseudo-labeled samples and lower thresholds introduce noisy labels, the confidence level $\tau = 0.7$ was empirically selected to balance pseudo-label number and quality. 

\subsection{Statistical Analysis}

We perform paired two-tailed t-tests comparing supervised and SSL accuracies across the five cross-validation folds for each (classifier, labeled ratio) pair in order to determine whether observed SSL improvements are statistically significant.

The null hypothesis $H_0$ states that SSL provides no improvement over supervised learning (mean difference $\mu_d = 0$). For a given classifier and labeled ratio, let $d_1, d_2, \ldots, d_5$ be the per-fold differences (SSL accuracy minus supervised accuracy). The t-statistic is computed as:

\begin{equation}
t = \frac{\bar{d}}{s_d / \sqrt{5}}
\end{equation}

where $\bar{d}$ is the mean difference across folds and $s_d$ is the sample standard deviation of the differences. We reject $H_0$ when the two-tailed p-value $p < 0.05$, indicating a statistically significant improvement (positive $\bar{d}$) or degradation (negative $\bar{d}$). All statistical tests are implemented using \texttt{scipy.stats.ttest\_rel}.

\subsection{Upper Bound Estimation}

To determine the best possible performance for each classifier, we train a fully supervised model on 100\% of the training pool labels (8118 samples) and assess it on the held-out test set (3480 samples). This provides an upper bound that can be used to compare SSL performance. The premise is that the method is practically useful for lowering labeling costs if SSL with limited labels approaches the upper bound. 

\begin{figure}[!t]
\centering
\includegraphics[width=\columnwidth]{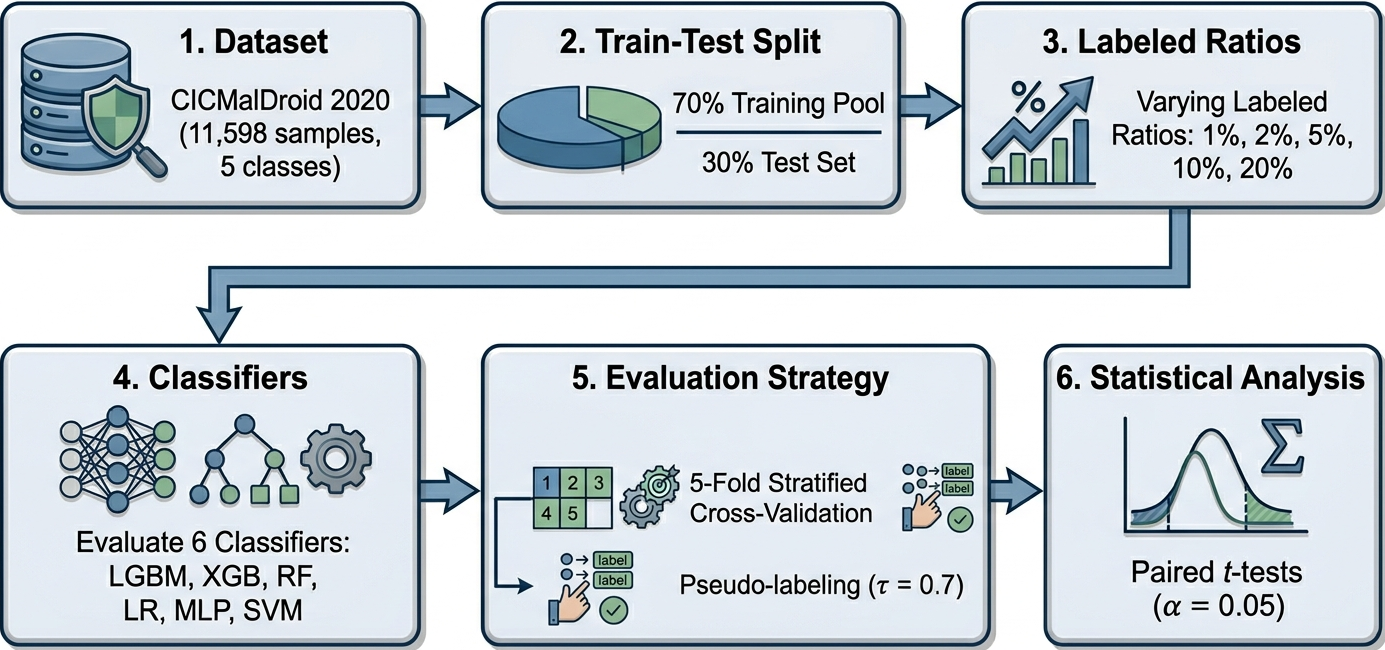}
\caption{Summary of the experimental methodology.}
\label{fig:pipeline}
\end{figure}

Figure~\ref{fig:pipeline} provides an overview of the experimental pipeline.

\section{Experimental Results and Analysis}

\subsection{Upper Bound Performance}

Table~\ref{tab:upper-bound} reports the fully supervised performance of all six classifiers trained on 100\% of the training labels (8118 samples) and evaluated on the held-out test set (3480 samples). LGBM achieves the highest accuracy (94.66\%) and macro F1 (93.61\%), followed closely by XGB (94.34\%) and RF (94.08\%). LR and MLP lag behind, while due to SVM's sensitivity to feature scaling and parameter selection under default settings, it exhibits the lowest performance (77.93\%).

\begin{table}[htbp]
\caption{Fully supervised upper bound performance (100\% labels).}
\begin{center}
\begin{tabular}{l|c|c}
\hline
\textbf{Classifier} & \textbf{Test Accuracy} & \textbf{Test Macro F1} \\
\hline
LightGBM (LGBM) & 94.66\% & 93.61\% \\
XGBoost (XGB) & 94.34\% & 93.31\% \\
Random Forest (RF) & 94.08\% & 92.87\% \\
MLP & 90.29\% & 88.39\% \\
Logistic Regression (LR) & 84.77\% & 82.55\% \\
SVM  & 77.93\% & 74.38\% \\
\hline
\end{tabular}
\label{tab:upper-bound}
\end{center}
\end{table}

Each classifier's performance ceiling is determined by these upper bounds. Notably, at higher label ratios, LGBM, XGB, and RF perform perfectly well ($>$94\% accuracy), and that leaves little opportunity for SSL improvement.

\subsection{SSL Benefit Summary}

Table~\ref{tab:ssl-improvement} presents the SSL improvement (SSL accuracy $-$ supervised accuracy) for each classifier at five labeled ratios ranging from 1\% to 20\%. Values correspond to mean $\pm$ standard deviation over five-fold cross-validation. Statistically significant improvements at $p < 0.05$ are shown in bold with asterisks (*), whereas significant degradations are in bold with daggers (†). The results reveal three distinct patterns:

\begin{table}[htbp]
\caption{SSL improvement over supervised baseline (gap in percentage points).}
\begin{center}
\begin{tabular}{|p{33pt}|p{25pt}|p{28pt}|p{28pt}|p{28pt}|p{28pt}|}
\hline
\textbf{Classifier} & \textbf{1\%} & \textbf{2\%} & \textbf{5\%} & \textbf{10\%} & \textbf{20\%} \\
 & & \textbf{Labels} & \textbf{Labels} & \textbf{Labels} & \textbf{Labels} \\
\hline
LGBM & $+0.7$ & $\mathbf{+1.3^*}$ & $\mathbf{+0.8^*}$ & $+0.3$ & $-0.1$ \\
 & $\pm 1.8$ & $\pm 1.3$ & $\pm 1.3$ & $\pm 1.0$ & $\pm 0.9$ \\
\hline
XGB & $+0.3$ & $0.0$ & $+0.3$ & $+0.5$ & $+0.3$ \\
 & $\pm 1.7$ & $\pm 1.9$ & $\pm 1.2$ & $\pm 1.0$ & $\pm 0.7$ \\
\hline
RF & $\mathbf{-3.1^\dagger}$ & $\mathbf{-1.6^\dagger}$ & $-0.9$ & $-0.5$ & $+0.1$ \\
 & $\pm 4.9$ & $\pm 2.5$ & $\pm 1.3$ & $\pm 0.9$ & $\pm 0.7$ \\
\hline
LR & $-0.9$ & $-0.6$ & $\mathbf{-0.4^*}$ & $-0.3$ & $-0.4$ \\
 & $\pm 1.3$ & $\pm 1.5$ & $\pm 1.2$ & $\pm 0.6$ & $\pm 0.8$ \\
\hline
MLP & $-4.3$ & $+2.4$ & $+3.3$ & $+1.4$ & $+1.3$ \\
 & $\pm 5.9$ & $\pm 4.5$ & $\pm 4.4$ & $\pm 2.6$ & $\pm 1.9$ \\
\hline
SVM & $0.0$ & $+0.4$ & $\mathbf{+4.4^*}$ & $+0.5$ & $+0.2$ \\
 & $\pm 2.7$ & $\pm 2.9$ & $\pm 3.0$ & $\pm 1.6$ & $\pm 1.0$ \\
\hline
\multicolumn{6}{l}{\footnotesize $^*$ $p < 0.05$ (statistically significant improvement)} \\
\multicolumn{6}{l}{\footnotesize $^\dagger$ $p < 0.05$ (statistically significant degradation)} \\
\end{tabular}
\label{tab:ssl-improvement}
\end{center}
\end{table}

\paragraph{Classifiers That Benefit from SSL}

SVM shows the largest statistically significant gain (+4.4\% at 5\% labels, $p = 0.0028$), which indicates that weaker classifiers with higher bias benefit more from pseudo-labeled data. LGBM also gets minor gains at 2\% (+1.3\%, $p = 0.0015$) and 5\% (+0.8\%, $p = 0.0275$).

\paragraph{Classifiers Harmed by SSL}

RF shows continuous degradation with pseudo-labeling, and there are significant performance drops at 1\% ($-3.1\%$, $p = 0.0048$) and 2\% ($-1.6\%$, $p = 0.0268$). This suggests that the bagging ensemble's inherent diversity may be disrupted by noisy pseudo-labels.

\paragraph{Classifiers with No Significant SSL Effect}

XGB shows no statistically significant change at any labeled ratio. MLP shows promising positive gaps (+2.4\% at 2\%, +3.3\% at 5\%) but high variance prevents statistical significance ($p > 0.07$). LR shows either no effect or small negative effects.

We analyze per-class F1 scores for SVM at 5\% labels, the configuration showing the strongest overall SSL improvement, to understand which malware families benefit most from SSL. Table~\ref{tab:per-class-svm} presents the per-class F1 scores across five-fold cross-validation.

\begin{table}[htbp]
\caption{Per-class F1 scores for SVM at 5\% labels.}
\begin{center}
\begin{tabular}{|p{48pt}|c|c|c|}
\hline
\textbf{Class} & \textbf{Supervised F1 (\%)} & \textbf{SSL F1 (\%)} & \textbf{Change (pp)} \\
\hline
Adware & $12.5 \pm 7.5$ & $26.3 \pm 8.4$ & $\mathbf{+13.8}$ \\
\hline
Banking & $36.5 \pm 7.5$ & $44.3 \pm 5.8$ & $\mathbf{+7.8}$ \\
\hline
SMS Malware & $70.7 \pm 3.6$ & $76.3 \pm 2.1$ & $\mathbf{+5.6}$ \\
\hline
Riskware & $53.7 \pm 4.6$ & $57.7 \pm 3.7$ & $\mathbf{+3.9}$ \\
\hline
Benign & $62.8 \pm 2.9$ & $63.6 \pm 2.1$ & $+0.8$ \\
\hline
\end{tabular}
\label{tab:per-class-svm}
\end{center}
\end{table}

SSL provides the largest gains for the classes that are most difficult to classify under supervised learning. Adware, which achieves only 12.5\% F1 with supervised learning (barely above random chance for five classes), improves by a dramatic +13.8 percentage points with pseudo-labeling. Banking, the second hardest class (36.5\% F1 supervised), improves by +7.8\%. In contrast, Benign, which already achieves 62.8\% F1 supervised, sees negligible improvement (+0.8\%). Adware remains a challenging class even with SSL, but the improvement from $12.5\%$ to $26.3\%$ F1 suggests that pseudo-labeling helps the model better distinguish this category.

\subsection{Effect of Labeled Data Quantity}

\begin{figure*}[!t]
\centering
\includegraphics[width=0.9\textwidth]{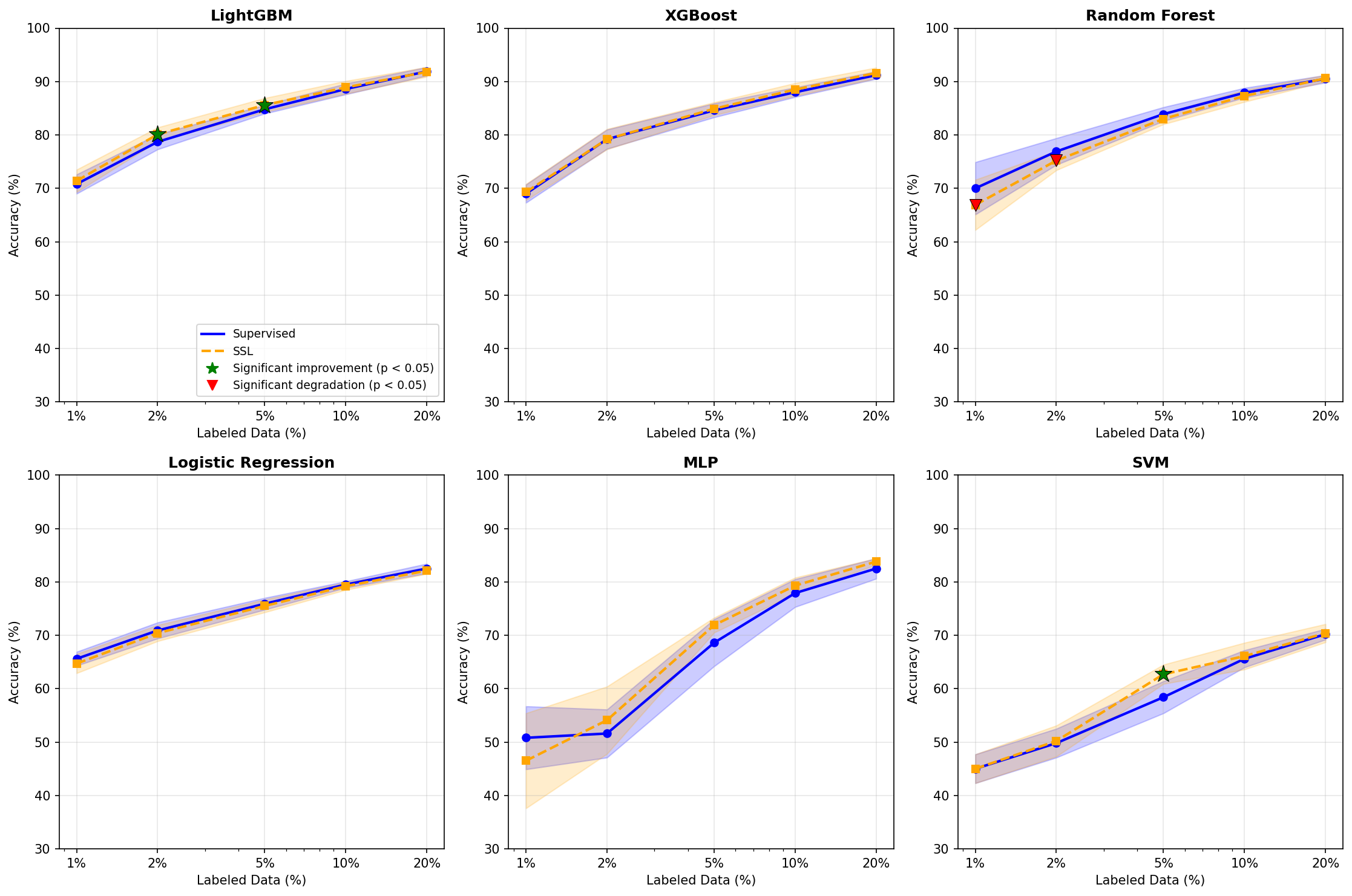}
\caption{Accuracy vs. labeled data ratio for all six classifiers. Solid blue lines represent supervised learning; dashed orange lines represent SSL with pseudo-labeling. Shaded regions indicate $\pm 1$ standard deviation across five-fold cross-validation. SSL shows clear benefit for SVM at 5\% labels and modest benefit for LGBM at 2--5\% labels, while degrading RF at low label ratios.}
\label{fig:figure1}
\end{figure*}

Figure \ref{fig:figure1} reveals three key insights:

\paragraph{Diminishing Returns Beyond 10\% Labels}
All classifiers reach within 2--3\% of their upper bound accuracy with only 10\% labeled data (around 811 samples). This suggests that for Android malware classification on this dataset, a collection of approximately 800 representative samples produces near-optimal performance.

\paragraph{SSL's Benefit Concentrates at Low Label Ratios}
The largest SSL improvements seem to occur at 1--5\% labels, where labeled data is most scarce. At 20\% labels, no classifier shows statistically significant SSL benefit, as supervised learning already approaches the upper bound.

\paragraph{High Variance for Neural Networks}
MLP's sensitivity to random initialization and data sampling is demonstrated by the highest standard deviations over the folds (up to $\pm5.9\%$ at 1\% labels). The observed $+3.3\%$ gain at 5\% labels does not achieve statistical significance, which can be explained by this instability.

\subsection{Upper Bound Gap Analysis}

Table~\ref{tab:upper-gap} reports the gap between each SSL configuration and the fully supervised upper bound (from Table~\ref{tab:upper-bound}). For example, at 5\% labels, LGBM with SSL achieves 85.6\% accuracy, compared to its upper bound of 94.66\%, indicating a gap of 9.06 percentage points.

\begin{table}[htbp]
\caption{Gap to upper bound under SSL (percentage points).}
\begin{center}
\begin{tabular}{|p{33pt}|p{28pt}|p{28pt}|p{28pt}|p{28pt}|p{28pt}|}
\hline
\textbf{Classifier} & \textbf{1\%} & \textbf{2\%} & \textbf{5\%} & \textbf{10\%} & \textbf{20\%} \\
 & & \textbf{Labels} & \textbf{Labels} & \textbf{Labels} & \textbf{Labels} \\
\hline
LGBM & 23.3 & 14.6 & 9.1 & 5.8 & 2.8 \\
\hline
XGB & 25.1 & 15.2 & 9.5 & 5.9 & 2.7 \\
\hline
RF & 27.2 & 18.9 & 11.1 & 6.8 & 3.5 \\
\hline
LR & 20.1 & 14.4 & 9.3 & 5.6 & 2.7 \\
\hline
MLP & 44.1 & 36.6 & 18.6 & 11.0 & 6.5 \\
\hline
SVM & 32.9 & 27.7 & 15.2 & 11.8 & 7.6 \\
\hline
\end{tabular}
\label{tab:upper-gap}
\end{center}
\end{table}

The gap analysis reveals two important insights. First, simple models with low upper bounds tend to benefit more from SSL in relative terms since SVM closes its upper bound gap from 32.9\% (supervised at 1\%) to 15.2\% (SSL at 5\%). Second, diminishing returns occur early, as moving from 5\% to 10\% labeled data reduces the gap by approximately 3 to 6 percentage points across all classifiers, whereas moving from 10\% to 20\% labeled data produces only 2 to 4 additional percentage points of improvement.

\subsection{Key Findings}

The experimental results lead to the following five key conclusions:

\textbf{Classifier-dependent SSL benefit:} Pseudo-labeling significantly improves SVM ($+4.4\%$ at 5\% labels) and modestly improves LGBM ($+0.8\text{--}1.3\%$ at 2--5\% labels), but significantly degrades RF ($-3.1\%$ at 1\% labels, $-1.6\%$ at 2\% labels).

\textbf{SSL helps only when labels are scarce:} None of the classifiers show statistically significant SSL improvement at 20\% labels, suggesting that collecting moderately labeled datasets (approximately 1600 samples) obviates the need for SSL.

\textbf{Upper bound gaps close rapidly:} All classifiers reach within 6–12\% of their fully supervised upper bound under SSL, with only 10\% labels (approximately 811 samples), and that is why extensive labeling may not be required for practical deployment.

\textbf{Tree-based methods are label-efficient:} LGBM and XGB achieve $>$90\% accuracy with only 10\% labels, making them strong candidates for malware detection in low-label scenarios without requiring SSL.

\textbf{SSL disproportionately benefits hard-to-classify families:} Based on per-class analysis of SVM at 5\% labels, Adware ($+13.8\%$, from $12.5\%$ to $26.3\%$ F1) and Banking ($+7.8\%$, from $36.5\%$ to $44.3\%$ F1), the classes with the lowest supervised performance receive the most SSL benefits.

\section{Discussion}

The per-class analysis reveals that Adware and Banking have the lowest supervised F1 (12.5\% and 36.5\%) but receive the largest gains ($+13.8\%$ and $+7.8\%$, respectively), and it can be attributed to SVM's margin sensitivity. The decision boundary is poorly estimated with limited labeled data, and it is specifically applicable for classes that have overlapping feature distributions, like Adware and Benign applications. In addition to providing extra support vectors that help to fine-tune the boundary, pseudo-labeled samples effectively "fill in" the feature space. The strong inverse correlation between supervised F1 and SSL gain suggests that pseudo-labeling is most valuable when the initial decision boundary is poorest.

RF shows statistically significant degradation at 1\% ($-3.1\%$) and 2\% ($-1.6\%$) labels. Bagging already provides ensemble diversity that mitigates high variance. Adding pseudo-labels introduces correlated noise across trees and reduces effective diversity. RF probability estimates are not as well-calibrated compared to estimates from boosting techniques. This means the confidence threshold $\tau = 0.7$ may select pseudo-labels that are confidently wrong. Therefore, practitioners should not assume that SSL can be advantageous for all ensemble methods.

MLP shows promising absolute increases ($+2.4\%$ at 2\%, $+3.3\%$ at 5\% labels), but its high cross-validation variance ($\sigma > 4\%$ at low label ratios) prevents it from reaching statistical significance. The confidence threshold is unreliable because of sensitivity to random weight initialization, optimization difficulties with sparse labeled data, and overconfident probability outputs. The constant positive gap at 2–5\% labels indicates that MLP would benefit from SSL with more regularization or better architectural choices, even though it is not statistically significant.

LGBM improves modestly at 2\% ($+1.3\%$) and 5\% ($+0.8\%$) labels, but gains disappear at higher ratios. Tree-based methods are inherently label-efficient. LGBM achieves 90\% accuracy with only 10\% labels without any SSL. The small SSL gain suggests that LGBM's built-in regularization already provides much of the benefit that pseudo-labeling offers. Practitioners using LGBM may find that simply collecting more labeled data (to 10\% of the total, approximately 811 samples) is more cost-effective than implementing SSL.

\section{Conclusion}

This paper introduced a systematic assessment of pseudo-labeling for semi-supervised Android malware classification over the CICMalDroid 2020 dataset. For each of the six classifiers, across five labeled ratios (1\% to 20\%), we evaluated whether there is a statistically significant difference in performance between SSL and supervised learning.

Our experiments show that the benefits from SSL is classifier-dependent. SVM shows the largest and most significant gain ($+4.4\%$ at 5\% labels, $p = 0.0028$). The hardest-to-classify families, Adware ($+13.8\%$) and Banking ($+7.8\%$), gain the most improvements, according to per-class analysis. LGBM shows modest but notable gains at 2--5\% labels ($+0.8\text{--}1.3\%$). A notabale finding is that RF drastically deteriorates at low label ratios ($-3.1\%$ at 1\% and $-1.6\%$ at 2\%). At the same time, MLP displays encouraging absolute increases ($+2.4\text{--}3.3\%$) that fall short of statistical significance because of high variance.

There are a number of limitations to this study. We only looked at pseudo-labeling, leaving more advanced SSL techniques (such as MixMatch, UDA, and FixMatch) for further investigation. The default hyperparameters were utilized by all classifiers, and adjusting them for each label ratio may produce different results. The findings might not apply to more recent Android versions or other malware datasets (Drebin, AndroZoo). Future research should investigate cross-dataset generalization, active learning combinations, adaptive confidence thresholds, and consistency regularization for neural networks.

Despite these limitations, our findings serve as the first systematic, classifier-dependent evaluation of SSL for Android malware classification. These results provide researchers and practitioners with guidance to decide whether pseudo-labeling might be a worthwhile strategy and with which classifier.

\end{document}